\documentclass{article}

\usepackage{graphicx} 
\usepackage{subcaption} 

\usepackage{natbib}

\usepackage{algorithm}
\usepackage{algorithmic}
\usepackage{hyperref}
\usepackage{booktabs}
\usepackage{multirow}
\usepackage{amsmath}

\newif\ifshowcomments
\showcommentstrue
\ifshowcomments
\newcommand\jirvin[1]{\textcolor{red}{[jirvin: #1]}}
\newcommand\pranav[1]{\textcolor{blue}{[pranav: #1]}}
\else
\newcommand\jirvin[1]{}
\newcommand\pranav[1]{}
\fi
\usepackage{array}

\usepackage[accepted]{icml2017}

\icmltitlerunning{Scene Learning: DCN For Wind  Power Prediction by Embedding Turbines into Grid Space}

\begin{document} 

\twocolumn[
\icmltitle{Scene Learning: Deep Convolutional Networks For Wind \\ Power Prediction by Embedding Turbines into Grid Space}

\icmlsetsymbol{equal}{*}

\begin{icmlauthorlist}
\icmlauthor{Ruiguo Yu}{cs}
\icmlauthor{Zhiqiang Liu}{cs}
\icmlauthor{Xuewei Li}{cs}
\icmlauthor{Wenhuan Lu}{scs}
\icmlauthor{Mei Yu}{cs}
\icmlauthor{Jianrong Wang}{cs}
\icmlauthor{Bin Li}{eie}
\end{icmlauthorlist}

\icmlaffiliation{cs}{School of Computer Science and Technology, Division of Intelligence and Computing, Tianjin University}
\icmlaffiliation{scs}{School of Computer Software, Division of Intelligence and Computing, Tianjin University}
\icmlaffiliation{eie}{School of Electrical and Information Engineering , Tianjin University}

\icmlcorrespondingauthor{Mei Yu}{yumei@tju.edu.cn}


\icmlkeywords{Wind, Embedding, Grid Space, Prediction, Convolutional Networks }
\vskip 0.1in

]

\printAffiliationsAndNotice{}

\begin{abstract} 
Wind power prediction is of vital importance in wind power utilization. There have been a lot of researches based on the time series of the wind power or speed, but In fact, these time series cannot express the temporal and spatial changes of wind, which fundamentally hinders the advance of wind power prediction. In this paper, a new kind of feature that can describe the process of temporal and spatial variation is proposed, namely, Spatio-Temporal Features. We first map the data collected at each moment from the wind turbine to the plane to form the state map, namely, the scene, according to the relative positions. The scene time series over a period of time is a multi-channel image, i.e. the Spatio-Temporal Features. Based on the Spatio-Temporal Features, the deep convolutional network is applied to predict the wind power, achieving a far better accuracy than the existing methods. Compared with the starge-of-the-art method, the mean-square error ($MSE$) in our method is reduced by 49.83\%, and the average time cost for training models can be shortened by a factor of more than 150.

\end{abstract}


\section{Introduction}
Wind power has become a significant renewable resource that can be developed and utilized on a large scale. With the mass production of equipment, wind power has turned to be the fastest growing renewable energy in the world. By 2017, the worldwide wind power installed capacity has reached 539 GW, and 52GW was added in 2017 \cite{Windstatus2017}, thus making wind power expected to be one of the major power sources in the 21st century. However, due to the influence of wind speed and direction, randomness and volatility of wind turbines can not be avoided, bringing severe challenges to the safety and stability of the operation of power systems. Accurate wind power prediction can enhance the controllability of wind power, ensure the stable operation of the power grid, and promote the ability of the grid to accept wind power.

 At present, scholars have done a lot of related researches, including physical methods, statistical methods and machine learning methods. Among them, machine learning methods, including support vector machine regression (SVR) \cite{Chen2014Short}, k-nearest neighbor regression (kNN) \cite{Becker2017Completion} or multi-layer perceptron neural network (MLP) \cite{Deo2018Multi} are used to model wind speed time series or power time series to achieve prediction. Machine learning methods simplify the wind power forecasting problem, but it is difficult to improve the accuracy rate in recent years.

We think that the wind is temporal and spatial correlation process, however, the time series can only express the information at the time level, but say nothing at the space level, let alone the spatio-temporal process of air flow, thus fundamentally standing in the way of the progress of wind power prediction. Therefore, finding the features that can better express the state of the wind farm is the key to breaking through the bottleneck of accuracy.

Such being the case, this paper put forward a new feature that can express the spatio-temporal process of air flow, called spatio-temporal feature (STF).
The scene time series over a period of time is a multi-channel image, in which each scene is a sample of the true distribution of physical data in space, expressing spatial-related information. The scene sequence represents the change of wind farm state over time, expressing time information, so the multi-channel image is called spatio-temporal feature. Compared with wind speed or power series, STF implies factors such as wind speed, wind direction and air density, which greatly expands the ability to express wind-related information and lays a foundation for breaking through the bottleneck of wind power prediction accuracy.

Based on the STF, the spatio-temporal process of the wind farm is simulated and predicted by using the deep convolutional network, which has achieved good effects. The experimental results of 592 wind turbines in a certain area show that, the method proposed by us is better than exist stage-of-the-art series modeling methods, for the reason that the $MSE$ of the proposed method decreases by an average of 26.69\% and decreases by 49.83\% at most, and the time for training models is shortened by more than 150 times.

The innovations of this paper are as follows:

\begin{itemize}
\item {The spatio-temporal feature in the form of the multichannel image is constructed by embedding the grid space of the wind motor, which fully expresses the spatio-temporal variation process of the air flow and can perfectly combine with the most advanced theory of deep learning at present.}
\item {Two kinds of deep convolutional network models that are suitable to use the spatio-temporal feature for wind power prediction are proposed, which can predict the wind power of a large number of turbines in parallel. And the accuracy and time cost of the prediction have been greatly optimized.}
\end{itemize}

\section{Related Work}

\subsection{Machine Learning Methods in WPP}
The machine learning method performs well in short term prediction. By means of the regression model or neural network, researchers map the time series to the wind power of the future moment, so as to make the prediction. The commonly used methods are SVR \cite{Chen2014Short}, kNN \cite{Becker2017Completion}, Multilayer Perceptron Network (MLP) \cite{Deo2018Multi, Marvuglia2012Monitoring} and Long and Short Term Memory Neural Network ( LSTM ) \cite{Qu2016Short}, etc., among which SVR and kNN are the representatives.

SVR has a perfect mathematical foundation in theory and performs best among numerous regressions. It realizes regression by finding a hyperplane to make all the data closest to the plane. This process can be abstracted as that, when equation \eqref{eq:svr_condition} is satisfied, the parameter should be found to make the value of equation \eqref{eq:svr_target} minimized. In these two equations,  $C$ and $\varepsilon$ are empirical parameters, $\xi_i$, and $\xi_i^*$ are called relaxation factors, and $w$ and $b$ represent hyperplanes \cite{Treiber2016Wind}.

\begin{equation}
\frac{1}{2} \parallel w \parallel^2 + C\sum_{i=1}^n(\xi_i+\xi_i^*)
\label{eq:svr_target}
\end{equation}

\begin{equation}
\left\{
\begin{aligned}
y_i - \langle w, X_i \rangle -b & \le \varepsilon + \xi_i \\
\langle w, X_i \rangle + b - y_i & \le \varepsilon + \xi_i^* \\
\xi_i, \xi_i* & \ge 0
\end{aligned}
\right.
\label{eq:svr_condition}
\end{equation}

It has been proved in many literatures \cite{Treiber2016Wind,Chen2014Short,Khosravi2018Time,D2017Performance} that SVR is one of the best methods in the field of wind power prediction currently.

KNN is the most simply equipped machine learning model based on similarity metric and it still has a good performance in practice. As the similarity between the two vectors is negative correlated with the distance between them (such as Euclidean distance, Manhattan distance, etc.), the similarity can be represented by distance. The $k$ vectors in a set with the minimum distance from the target vector $x$ can be called $k$ nearest neighbors of $x$. If $\aleph_k(x)$ denotes the subscript of $k$ nearest neighbors of $x$ in the training set, then the prediction result $p(x)$ of kNN model for $x$ is generated by equation \eqref{eq:knn_predict} , and the function $f$ can use arithmetic average method, weighted average method or other methods that are more complex .

\begin{equation}
p(x)=f_{i \in \aleph_k(x)} (yi)
\label{eq:knn_predict}
\end{equation}

Based on the algorithms such as k-d tree, a kNN model can be trained very quickly. 

In recent years, there have been some new ideas in this research field. For example, in literatures \cite{Tascikaraoglu2016Exploiting, Lixin2018Prediction, Saroha2018Wind}, wavelet transform was used to decompose the power series to form multiple new sub-series that would be predicted in turn. And then the results were combined. This method needs to build a model for each sub-series, thereby leading to much higher costs. The researchers modeled the prediction error to improve the prediction effect by error analysis\cite{Wang2018A, Giorgi2011Error}. But the error is produced by the specific prediction model, which has limits, difficult to be applied to the production. In addition, the process of error analysis increases the computational costs. Although using ensemble learning to predict could improve the accuracy \cite{Zhao2016An, Pinson2013Skill,Wang2017Deep,Heinermann2016Machine}, many models working at the same time also consume computational resources substantially.  Moreover, in the literature \cite{Wang2017Deep}, the series data of at the length of p$\times$q were filled in the grid of p$\times$q in order to achieve a two-dimensional image, following which the convolutional neural network was utilized for the prediction. But the constructed image had no explicit physical meaning. And the required time series were far too long to add the computational costs.

In sum, all of the above approaches used the series data for modeling in essence, and achieved a higher accuracy via the complicated models. However, their computational cost was largely increased and their models could not reflect the spatio-temporal variation of air.

\subsection{Convolutional Neural Network}

This chapter introduces the convolutional neural network (CNN), which lays the foundation for the third chapter to introduce the method proposed in this paper. At present, CNN is the most successful method in deep learning that has been widely used in auxiliary medical treatment, speech recognition, intelligent city and automatic driving system. Besides, CNN can speed up computing by GPU. With the rapid development of hardware in recent years, the computing ability of computers has been greatly improved, thus leading the CNN model to a significant progress in many fields.

The central operation of CNN is the convolution. As both the input and output of convolutions are multichannel images, these images are usually called as feature maps. There are abundant types of CNN models, but as a whole they can be divided into two basic types. The first is the coding machine-decoder model, whose core operations are the convolution, pooling and deconvolution. The convolution process is to extract deep features, pooling is to narrow the size of images, and the deconvolution aims at enlarging the image size by up-sampling. FCN network \cite{Long2015Fully} is typical in this method. The second type is a convolutional network with a fully connected layer, whose the core operations include the convolution, pooling and full connection. In this type of model, the convolution and pooling process produce deep features, while full connection maps deep features to predictive values. On account of the excellent expression of full connection, the model, VGGNet \cite{Simonyan2014Very} in particular, can always fit a very complex nonlinear relation.

\begin{figure}[h]
  \centering
  \includegraphics[width=0.2\textwidth]{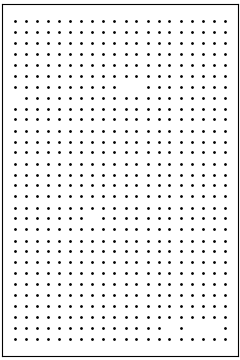}
  \caption{
       The image produced by scaling down the real coordinates. White pixels indicate blanks and black pixels indicate wind motors. Black pixels are extremely sparse, that is, the ratio of effective pixels in the picture is very low.
  }
  \label{fig:real_loc}
\end{figure}


\begin{figure}[ht!]
  \centering
  \includegraphics[width=8.2cm]{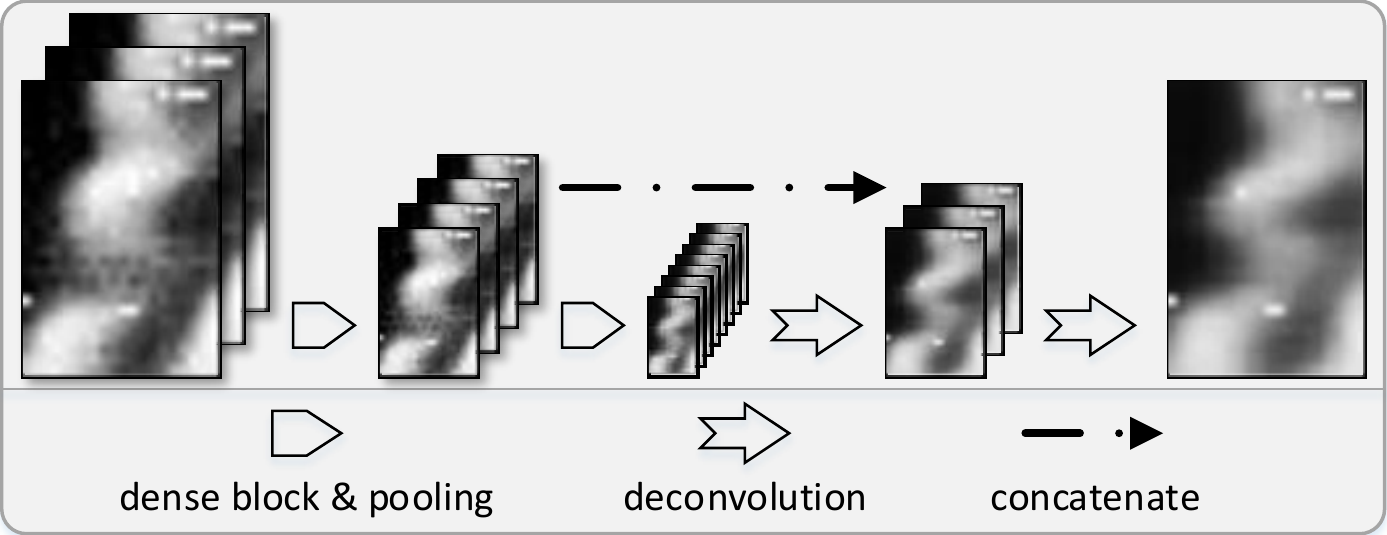}
  \caption{ Structure of the E2E model  }
  \label{fig:e2e}
\end{figure}

\begin{figure}[ht!]
  \centering
  \includegraphics[width=8.2cm]{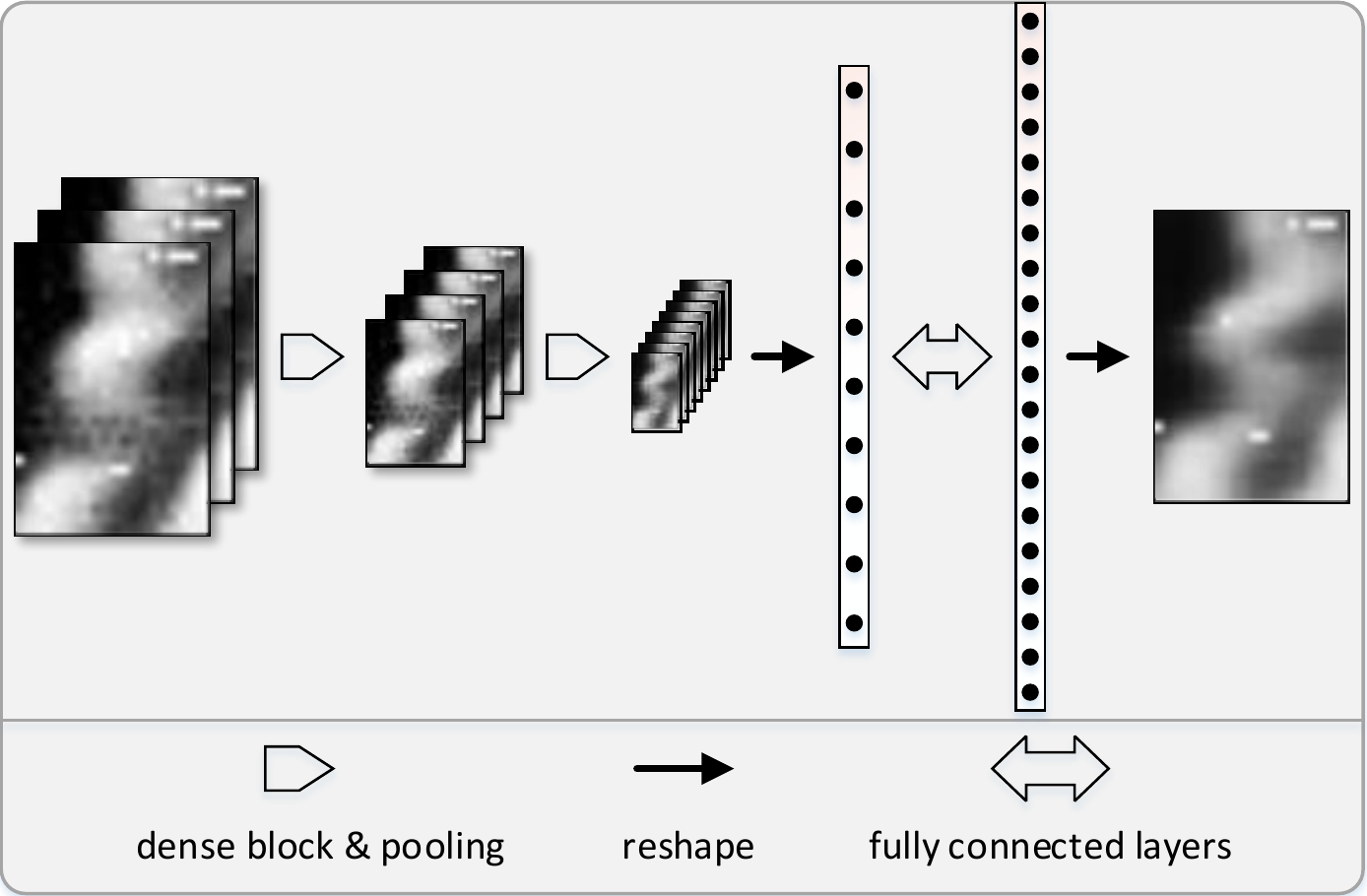}
  \caption{ Structure of the FC-CNN model  }
  \label{fig:fc-cnn}
\end{figure}

\section{Proposed Method}

The information related to wind power such as turbine’s output power and wind speed can be strongly combined with convolutional networks. On one hand, convolutional networks are quite suitable to deal with grid data structure, which can automatically extract features at different layers and realize the end-to-end learning. On the other hand, the wind turbine itself is in the grid space whose distribution is easily modeled as a planar grid structure. But the current researches have little experience in combining the both.

In this chapter, the spatio-temporal feature (STF) and its basis, scenes, are introduced, and then two kinds of convolutional networks models based on STF are put forward, which serve as two basic convolutional networks mentioned in the second chapter. In this paper, we use these two models in order to show that STF can be combined with various convolutional networks in practical utilization.

The rest of this chapter will further elaborated on the above contents.

\subsection{Scene and STF}

Feature extraction has always been a hot topic in wind power prediction. In this paper, the feature extracted only from the data of the target turbine itself is called “single-feature (SF)”, and the feature extracted from the data of the target turbine and several adjacent generators is called “local-feature (LF)”. Basically, the local-feature is an extended form of the single-feature. When the local-feature selects a distance threshold of 0 for adjacent turbines, it degenerates into single-feature.

Most of the features used in existing works are single-features, and some researchers have also studied local-features. For example, in the literature \cite{Treiber2016Wind}, the local-feature is generated by connecting the single-feature of each turbine. The feature extracted in this way contains more information, but these information is not efficient, covering only the concept of temporal level but in devoid of the spatial dimension. 


In order to describe the spatial distribution of wind in a certain area at a certain time, the concept of scene is set out in this paper. We map the output electric power of the wind turbines to the plane according to the geographical coordinates of the turbines at a certain time, to form a two-dimensional image called the scene.
Mapping the real coordinates to the plane is the main problem while constructing a scene. The most direct solution is to scale down the real geographic coordinates and then to draw them onto the plane, as shown in figure \ref{fig:real_loc}.

This method can successfully show the spatial position, but the size of the constructed image is relatively large containing only sparse effective pixels, which is not conducive to calculation. To solve this problem, this paper proposes a method to embed turbines into grids as small an area as possible, which is called grid space embedding method. In this algorithm, the longitude and latitude coordinates are firstly processed by ridding unbalance and discretization, in order to determine the shape of the scene and then generate the grids. After grid generation, each turbine is mapped to the corresponding grid in the order sorted by its horizontal and vertical coordinates. More details are shown in algorithm \ref{algorithm: embeding}. The output is the mapping matrix $G$ of turbines to grid points, each position serving as the serial number of the turbine. When the vacant position is filled with $-1$ to get the matrix $G$, the output of turbines at a certain time is filled into the matrix according to the position specified by $G$, and then the scene corresponding to the time can be obtained, which is shown in algorithm \ref{algoithm:scening}.

The proposed embedding algorithm uses the grids as small as possible to avoid invalid pixels, and the constructed scene is suitable for convolutional computation.

The scene represents the spatial distribution of wind power at a certain time. And  connecting several continuous scenes in series can convey the process of spatial state changing with time. Although the air motion is complicated, it still shows certain regularity on the whole, and the scene series can reflect this regularity to some extent. In this paper, the multichannel image got by the scenes arranged in time series is named as the spatio-temporal feature (STF).

\begin{algorithm}[h]  
\caption{Algorithm for embedding wind turbines into a grid} 
\label{algorithm: embeding} 
\begin{algorithmic}[1]
\REQUIRE $ids$: Id list of turbines, $coordinates$: Coordinate list of turbines
\ENSURE $G$: $ids$ embeding result 
	\STATE $latitudes = coordinates[:, 0]$
	\STATE $longitudes = coordinates[:, 1]$
	\STATE $latitudes = unique(latitudes)$, $longitudes = unique(longitudes)$  
	\STATE $sorted\_latitudes = sort(latitudes) $ //increasing order
	\STATE $sorted\_longitudes = sort(longitudes) $ //increasing order
	\STATE $array\_shape = [len(latitudes), len(longitudes]$
	\STATE $G=array(-1, shape = array\_shape)$
	\FOR {$id$, $(x, y)$ in enumerate($coordinates$)}
		\STATE $index_x = get\_index(sorted\_latitudes, x)$
		\STATE $index_y = get\_index(sorted\_longitudes, y)$       
		\STATE $G[index_x][index_y] = id$
	\ENDFOR
	\RETURN $G$
\end{algorithmic}
\end{algorithm} 

\begin{algorithm}[h]  
\caption{Algorithm for constructing scene} 
\label{algoithm:scening} 
\begin{algorithmic}[1]
\REQUIRE $G$: Output of embeding algorithm, $ids$: Id list of turbines, $I$: Information (e.g., output power) of the wind turbine at a certain time 
\ENSURE $S$: Scene 
	\STATE $S=array(0, shape = get\_shape\_of(G))$
    	\FOR {$x$ in range(len(G))}
    		\FOR {$y$ in range(len(G[0]))}
			\IF {$G[x][y] == -1$}
    				\STATE continue
    			\ELSE
    				\STATE $id = G[x][y]$
    				\STATE $S[x][y] = I[id]$
    			\ENDIF
		\ENDFOR
	\ENDFOR
	\RETURN $S$
\end{algorithmic} 
\end{algorithm} 

Each channel of STF independently represents spatial information, and the combination of the multichannel sorting represents temporal information. It is a kind of global-feature for it can synthetically deliver the information in a large geographical area and a long time range. In fact, each channel of the STF can also be used to represent different types of information, such as wind power output, wind speed, temperature and so on. The STF, which combines many kinds of data, is called MSTF.
The STF can be processed by deep convolutional neural network. Convolution neural network is the most mature theory of deep learning at present, which, given perfect tools and frameworks, can give full play to the advantages of new technologies such as GPU acceleration.

\subsection{E2E Model}
The first kind of convolutional neural network model for wind power prediction based on STF is introduced in this section, which is called E2E model, using the idea of autoEncoder \cite{Vincent2008Extracting}.

After received, the input image will be handled in two stages. The first stage is down-sampling, that is, the coding stage, in which the deep features are extracted step by step and the image size is shrunk by means of multiple nested convolution layers and a pooling layer. The second is up-sampling, that is, the decoding stage, which mainly includes deconvolutional layers. By deconvolution, the size of the feature map is initially increased, and finally the output of the same size as the input image is obtained. As a result, the pixels of the input image and the pixels of the output image can be corresponded one-to-one to realize the end-to-end mapping.

In the down-sampling stage, under the guidance of the idea of "short circuit" in DenseNet, the outputs of multiple prepositive convolutional layers are connected in series, and then input to the next convolutional layer to preserve the spatial information of the original input image. Since the major task of this stage is to fully extract features, the number of channels in the feature image increases rapidly. The main task of the upper sampling stage is the fusion of features in order to produce the output. In this stage, the outputs of each convolutional layer are no longer connected in series, and the output of each deconvolution reduces the channels. In this way the single channel image is finally output.
The structure of the E2E model is shown in Figure \ref{fig:e2e}.

\subsection{FC-CNN Model}

The second model is a convolutional neural network containing a fully connected layer, called FC-CNN. After receiving the input image, the model also performs the operations of two stages.The first stage is similar to the down-sampling stage of E2E model, but the deeper layers are in demand in FC-CNN and the size of feature map is smaller. The second stage is the fully connected network. The deep features are mapped to the output of each turbine by fitting the complex function relationship with the fully connected layer. The output vector length of the last full connected layer, equal to the number of pixels in the input image, is reshaped to be two dimensional, and mapped to the pixels of the input image one by one.
The down-sampling process of the model also incorporates the idea of Dense Net, and the  model structure is shown in figure \ref{fig:fc-cnn}.

\begin{figure*}[ht!]
\centering
\begin{subfigure}[t]{0.12\textwidth}
\includegraphics[width=0.95\textwidth]{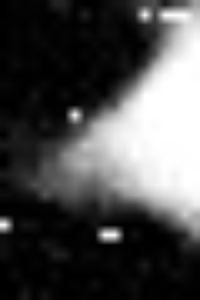}
\end{subfigure}%
\hfill
\begin{subfigure}[t]{0.125\textwidth}
\includegraphics[width=0.95\textwidth]{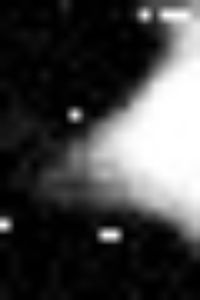}
\end{subfigure}%
\hfill
\begin{subfigure}[t]{0.125\textwidth}
\includegraphics[width=0.95\textwidth]{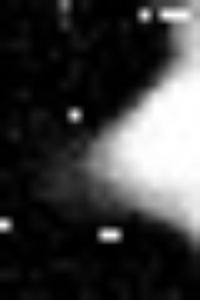}
\end{subfigure}%
\hfill
\begin{subfigure}[t]{0.125\textwidth}
\includegraphics[width=0.95\textwidth]{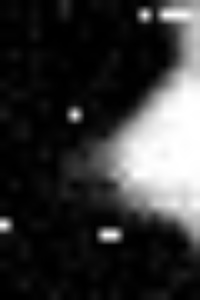}
\end{subfigure}%
\hfill
\begin{subfigure}[t]{0.125\textwidth}
\includegraphics[width=0.95\textwidth]{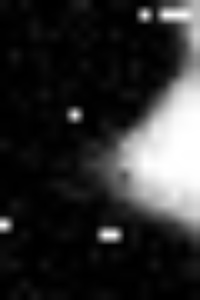}
\end{subfigure}%
\hfill
\begin{subfigure}[t]{0.125\textwidth}
\includegraphics[width=0.95\textwidth]{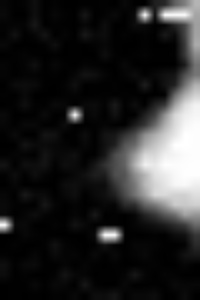}
\end{subfigure}%
\hfill
\begin{subfigure}[t]{0.12\textwidth}
\includegraphics[width=0.95\textwidth]{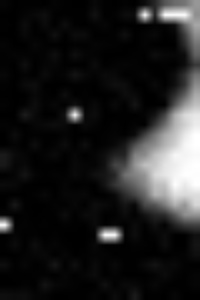}
\end{subfigure}%
\hfill
\begin{subfigure}[t]{0.125\textwidth}
\includegraphics[width=0.95\textwidth]{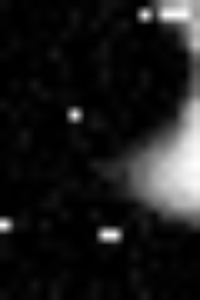}
\end{subfigure}%
\caption{Scene time series  related wind power.  The sequence involves 8 moments with a total duration of 70 minutes,  wherein the darker the pixels in the picture, the corresponding value of the wind power is larger.  The series clearly shows the state changes of the wind farm.}
\label{fig:scene_series}
\end{figure*}

\begin{table*}[t]
\caption{Comparison with existing methods on 592 wind turbines}
\label{tab:compare}
\centering
\begin{tabular}{cccccccccc} 
\toprule
  &   & SF+kNN & LF+kNN & SF+SVR & LF+SVR & STF+E2E & STF+FC-CNN & STF ensemble\\
\midrule
MSE ($MW^2$) 
     & MAX & 25.44 & 16.79 & 18.70 & 15.84 & \textbf{11.94} & \textbf{12.23} & \textbf{11.39}\\
     & MIN & 8.83  & 7.30  & 8.37  & 6.64  & \textbf{5.25}  & \textbf{5.00}  & \textbf{5.00}\\
     & AVE & 13.28 & 10.90 & 12.50 & 10.05 & \textbf{7.91}  & \textbf{7.78}  & \textbf{7.61}\\
\multicolumn{1}{c}{train time(s)}&- & \textbf{509} & 10592 & 91191 & 207081 & \textbf{1200} & \textbf{1059} & -\\
\bottomrule
\end{tabular}
\end{table*}

\begin{figure}[ht!]
  \centering
  \includegraphics[width=8.2cm]{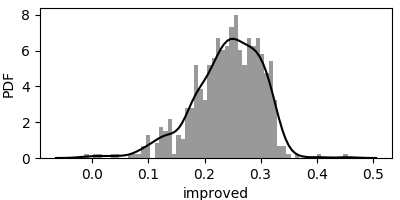}
  \caption{
       The ratio of $MSE$ reduction when FC-CNN compared to SVR using LF. The figure shows the results of experiments on 592 wind turbines. The abscissa represents the value calculated according to the equation \eqref{eq:each_turbine}, the vertical axis is the probability density, and the curve is the fitted probability density curve.
  }
  \label{fig:impro_svr}
\end{figure}

\begin{figure}[ht!]
  \centering
  \includegraphics[width=8.2cm]{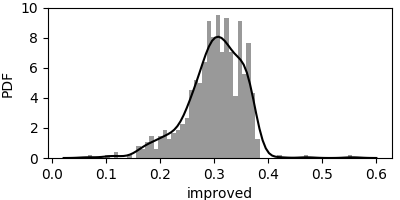}
  \caption{
       The ratio of $MSE$ reduction when FC-CNN compared to kNN using LF. The figure shows the results of experiments on 592 wind turbines. The abscissa represents the value calculated according to the equation \eqref{eq:each_turbine}, the vertical axis is the probability density, and the curve is the fitted probability density curve.
  }
  \label{fig:impro_knn}
\end{figure}

\begin{figure*}[ht!]
  \centering
  \includegraphics[width=0.9\textwidth]{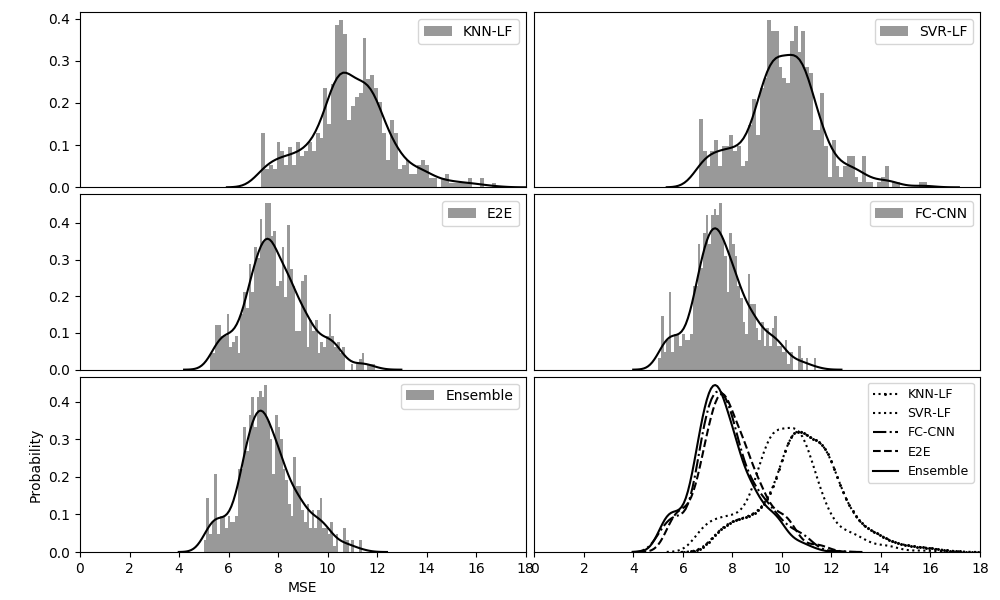}
  \caption{
       Predictive error distribution for each method
  }
  \label{fig:compare}
\end{figure*}

\begin{figure*}[ht!]
  \centering
  \includegraphics[width=0.9\textwidth]{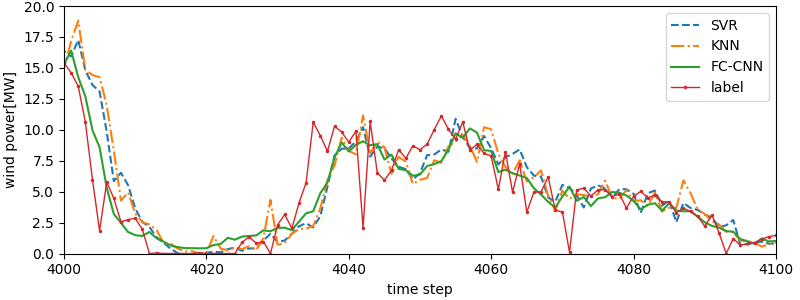}
  \caption{
       Comparison of the true wind power time series of a turbine and the predicted series corresponding to different methods.
  }
  \label{fig:pred_res}
\end{figure*}

\begin{figure*}[ht!]
\centering

\begin{subfigure}[t]{0.125\textwidth}
\includegraphics[width=0.95\textwidth]{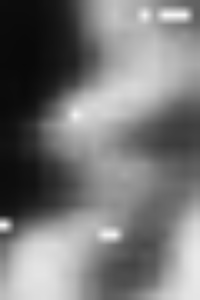}
\end{subfigure}%
\hfill
\begin{subfigure}[t]{0.125\textwidth}
\includegraphics[width=0.95\textwidth]{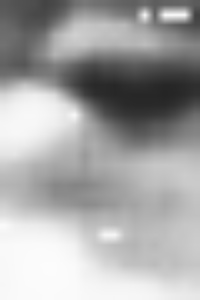}
\end{subfigure}%
\hfill
\begin{subfigure}[t]{0.125\textwidth}
\includegraphics[width=0.95\textwidth]{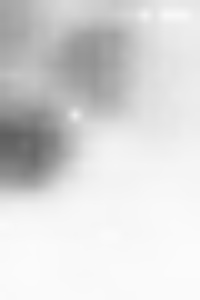}
\end{subfigure}%
\hfill
\begin{subfigure}[t]{0.125\textwidth}
\includegraphics[width=0.95\textwidth]{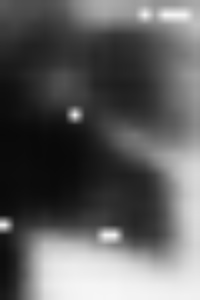}
\end{subfigure}%
\hfill
\begin{subfigure}[t]{0.125\textwidth}
\includegraphics[width=0.95\textwidth]{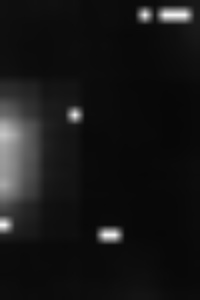}
\end{subfigure}%
\hfill
\begin{subfigure}[t]{0.125\textwidth}
\includegraphics[width=0.95\textwidth]{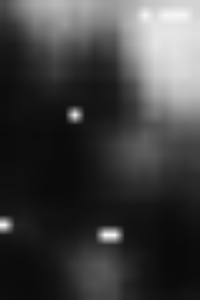}
\end{subfigure}%
\hfill
\begin{subfigure}[t]{0.125\textwidth}
\includegraphics[width=0.95\textwidth]{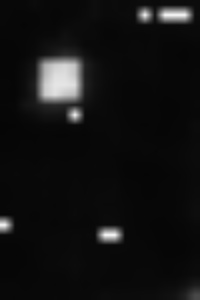}
\end{subfigure}%
\hfill
\begin{subfigure}[t]{0.125\textwidth}
\includegraphics[width=0.95\textwidth]{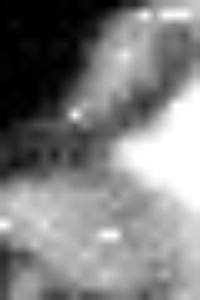}
\end{subfigure}%
\hfill

(a) Prediction Results of Wind Power

\begin{subfigure}[t]{0.125\textwidth}
\includegraphics[width=0.95\textwidth]{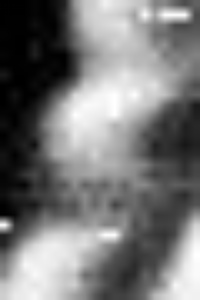}
\end{subfigure}%
\hfill
\begin{subfigure}[t]{0.125\textwidth}
\includegraphics[width=0.95\textwidth]{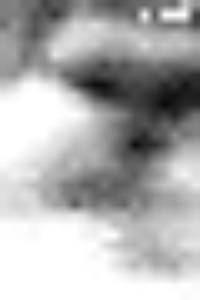}
\end{subfigure}%
\hfill
\begin{subfigure}[t]{0.125\textwidth}
\includegraphics[width=0.95\textwidth]{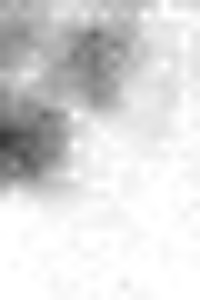}
\end{subfigure}%
\hfill
\begin{subfigure}[t]{0.125\textwidth}
\includegraphics[width=0.95\textwidth]{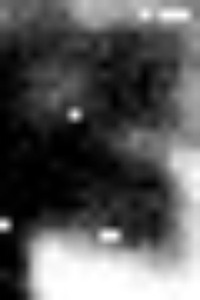}
\end{subfigure}%
\hfill
\begin{subfigure}[t]{0.125\textwidth}
\includegraphics[width=0.95\textwidth]{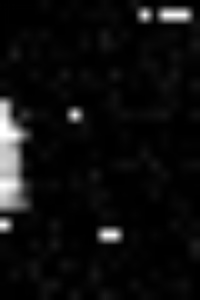}
\end{subfigure}%
\hfill
\begin{subfigure}[t]{0.125\textwidth}
\includegraphics[width=0.95\textwidth]{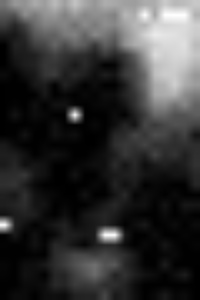}
\end{subfigure}%
\hfill
\begin{subfigure}[t]{0.125\textwidth}
\includegraphics[width=0.95\textwidth]{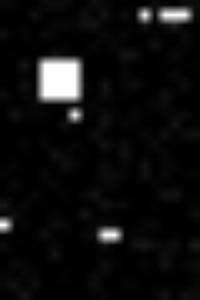}
\end{subfigure}%
\hfill
\begin{subfigure}[t]{0.125\textwidth}
\includegraphics[width=0.95\textwidth]{images/prediction_map_power/1450_ori.jpg}
\end{subfigure}%
\hfill

(b) Ground Truth of Wind Power

\begin{subfigure}[t]{0.125\textwidth}
\includegraphics[width=0.95\textwidth]{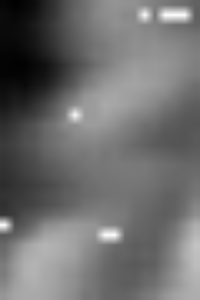}
\end{subfigure}%
\begin{subfigure}[t]{0.125\textwidth}
\includegraphics[width=0.95\textwidth]{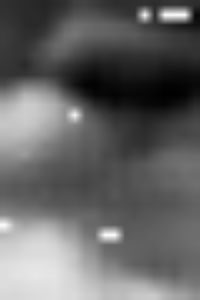}
\end{subfigure}%
\hfill
\begin{subfigure}[t]{0.125\textwidth}
\includegraphics[width=0.95\textwidth]{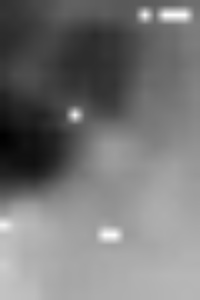}
\end{subfigure}%
\hfill
\begin{subfigure}[t]{0.125\textwidth}
\includegraphics[width=0.95\textwidth]{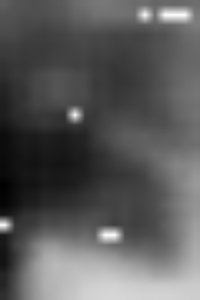}
\end{subfigure}%
\hfill
\begin{subfigure}[t]{0.125\textwidth}
\includegraphics[width=0.95\textwidth]{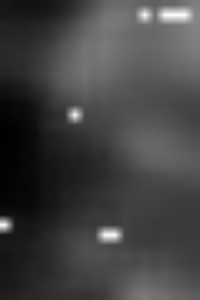}
\end{subfigure}%
\hfill
\begin{subfigure}[t]{0.125\textwidth}
\includegraphics[width=0.95\textwidth]{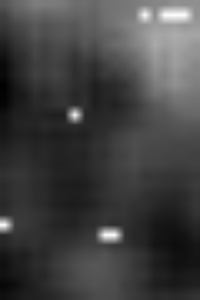}
\end{subfigure}%
\hfill
\begin{subfigure}[t]{0.125\textwidth}
\includegraphics[width=0.95\textwidth]{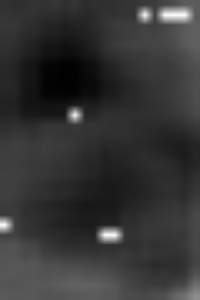}
\end{subfigure}%
\hfill
\begin{subfigure}[t]{0.125\textwidth}
\includegraphics[width=0.95\textwidth]{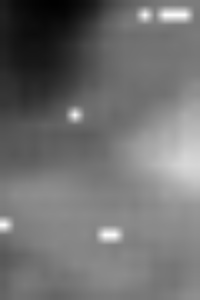}
\end{subfigure}%
\hfill

(c) Prediction Results of Wind Speed

\begin{subfigure}[t]{0.125\textwidth}
\includegraphics[width=0.95\textwidth]{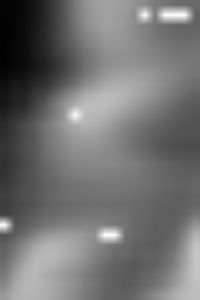}
\end{subfigure}%
\begin{subfigure}[t]{0.125\textwidth}
\includegraphics[width=0.95\textwidth]{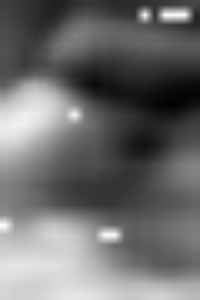}
\end{subfigure}%
\hfill
\begin{subfigure}[t]{0.125\textwidth}
\includegraphics[width=0.95\textwidth]{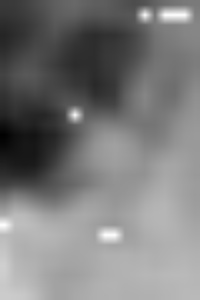}
\end{subfigure}%
\hfill
\begin{subfigure}[t]{0.125\textwidth}
\includegraphics[width=0.95\textwidth]{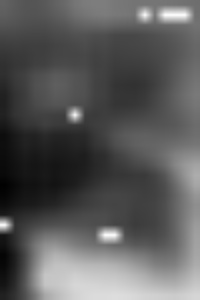}
\end{subfigure}%
\hfill
\begin{subfigure}[t]{0.125\textwidth}
\includegraphics[width=0.95\textwidth]{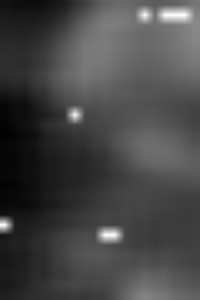}
\end{subfigure}%
\hfill
\begin{subfigure}[t]{0.125\textwidth}
\includegraphics[width=0.95\textwidth]{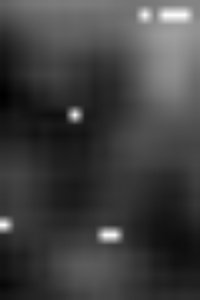}
\end{subfigure}%
\hfill
\begin{subfigure}[t]{0.125\textwidth}
\includegraphics[width=0.95\textwidth]{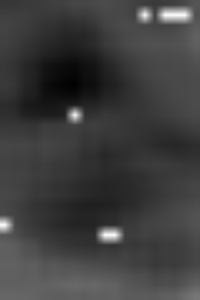}
\end{subfigure}%
\hfill
\begin{subfigure}[t]{0.125\textwidth}
\includegraphics[width=0.95\textwidth]{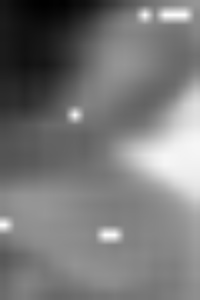}
\end{subfigure}%
\hfill
(d) Ground Truth of Wind Speed

\caption{ Scene predictions results corresponding to wind speed and wind power}
\label{fig:scene_compare}
\end{figure*}

\begin{table}
\label{tab:mstf1}
\caption{Comparison of effects between STF-based models and MSTF-based models}
\centering
\begin{tabular}{|c|c|c|c|c|} 
\hline
 Feature & \multicolumn{2}{|c|}{FC-CNN} &  \multicolumn{2}{c|}{E2E}\\ 
\hline
 Method  & \multicolumn{1}{|c|}{STF} &  \multicolumn{1}{|c|}{MSTF} & STF & MSTF \\
\hline
MAX  & 12.23 & 11.33 & 11.94 & 11.05 \\
MIN  & 5.00  & 4.72  & 5.25  & 4.96  \\
AVE  & 7.78  & 7.25  & 7.91  & 7.35  \\
\hline
\end{tabular}
\end{table}
\section{Experiment and Analysis}

\subsection{Data Sets and Evaluation Criteria}

The data set used here is the wind data set of the NREL 
{\footnote{\url{https://www.nrel.gov}}}
which contains the output values of every 10 minutes of wind turbines in the United States from 2004 to 2006. To validate our method, an area with the longitude range from 105.00W to 105.34W and a latitude range from 41.40N to 41.90N is selected, which is located in the middle of the United States, where wind turbines are densely distributed reaching a number of 592. And we make the prediction about the wind power output of the wind turbine after 30 minutes based on the above data. 

Accuracy is the most important factor to measure the effect of wind power prediction, and the main indexes of evaluating accuracy are mean square error ($MSE$) and root mean square error ($RMSE$), $RMSE$ being the square root of $MSE$. So in this paper, $MSE$ is chosen as the standard of evaluation, whose calculation process is shown in equation\eqref{eq:mse}, in which $real$ represent the series of true values, $predictions$ represent the series of the predicted values, and $n$ represents the length of the series.

\begin{equation}
MSE = \frac{1}{n}\sum^n_{i=0} {(real_i-predictions_i)^2}
\label{eq:mse}
\end{equation}

\subsection{Scene Display}

As shown in Figure \ref{fig:scene_series}, there are 8 scenes sequenced in time series, with the darker regions in each scene representing the larger value. This figure is used to show the spatial information expressed by scene and the spatio-temporal information expressed by STF.

It can be seen from the figure that the air flow in this region obviously shows regularity during this period (70min). Firstly, the output power of the wind turbine is strongly correlated with the spatial position. Secondly, as time goes by, a visible displacement is shown between the scenes. So it can be inferred that the west wind has crossed the border during this period, thus expanding the affected areas. These laws are the basis of prediction using machine learning methods. And the results show the advantage of STF, that is, being able to express the spatio-temporal variation process of wind. The traditional single-feature can be visualized into a curve, but it is difficult to find obvious regulation no matter for human eyes or computer algorithms, thus having no access to a better prediction accuracy.

\subsection{Experimental Results and Comparison}
The methods based on LF such as SVR have reached the level of stage-of-the-art in wind power prediction. In order to prove the validity of the method proposed in this paper, SVR which is the most accurate method for prediction is compared with kNN, the fastest training method. In the experiment, kNN uses SF training model when SVR uses LF training model. The experimental results are shown in tables \ref{tab:compare} and figure \ref{fig:compare}.

In the experiment, the $MSE$s of 592 wind turbines in each method are obtained firstly. table \ref{tab:compare} compares the effects of the methods according to the maximum, minimum and average values of these values. The average values of the $MSE$s in the two methods proposed in this paper are 7.91 and 7.78 respectively, and the integration of the two average values can reach the number of 7.61. However, the optimal value of the above standard of the existing methods is 10.05, compared with which that value in our methods is reduced by 24. 28\%. Therefore, according to the above numerical results, the two methods proposed in this paper are superior to other methods in prediction accuracy. 

Table \ref{tab:compare} provides a quantitative comparison of the overall performance of the methods. And figure \ref{tab:compare} further shows the distribution of each of these methods corresponding to $MSE$. The columnar section in each subgraph corresponds to the distribution of $MSE$, in which the curve illustrates the variation of probability density, the horizontal scale represents the value of $MSE$, and the ordinate represents the corresponding probability density ($PDF$). The first five images show the effect of each method, and the last image compares all the results, to find that the $MSE$ of FC-CNN and E2E model are distributed in the region with the smaller values. Therefore, on the whole, the proposed methods outperform the SVR and kNN.

The above results have proved the advantages of the proposed method. In figure \ref{fig:impro_svr} and figure \ref{fig:impro_knn}, wind turbines are analyzed in turn, to quantitatively compare the results of optimization. In the figures, $M$ denotes the models, LF+SVR and LF+kNN, used for comparison. The effect of the method using SF is inferior to that of the method using LC, so it is no longer comparison. The values got from equation \ref{eq:each_turbine} reflect the reduced ratio of $MSE$ of FC-CNN compared with M. And figure \ref{fig:impro_svr} and figure \ref{fig:impro_knn} are the probability density curves obtained by fitting these values.

\begin{equation}
p_i = \frac{MSE(M)_i-MSE(FC-CNN)_i}{MSE(M)_i}
\label{eq:each_turbine}
\end{equation}

In figure \ref{fig:impro_svr} and figure \ref{fig:impro_knn}, the area of the region whose horizontal coordinate is less than 0 is almost none, which means that the prediction effect of FC-CNN on almost all wind turbines is optimized compared with the above two methods. According to the statistics, compared with LC+SVR, its $MSE$ had an average reduction of 24.10\%, and maximally decreased by 45.55\%. And compared with LC+kNN, its $MSE$ decreased by 30.10\%, and highest by 45.55\%.

Figure 8 shows the predicted value curves of each method on a randomly selected turbine. It can be seen from the figure that the predicted results of the model using STF are more stable, whose stability is even better than that of the true value. As a matter of fact, wind is a natural phenomenon, but the conversion process from wind to wind power output is complex, with many interference factors related to the characteristics of the wind turbine itself. In order to further analyze the experimental results, the wind power prediction is divided into two stages. The first stage is to predict the information such as wind speed, and another stage is to convert the wind state information from the prediction to wind power output. And it has been believed in this paper that the prediction errors mainly come into existence during the second stage.
To verify this idea, this paper uses the proposed methods to separately predict wind speed and wind power. The typical results are shown in figure \ref{fig:scene_compare}, in which the true and predicted value of wind power and wind speed at 8 moments are visualized. Obviously, the predicted value of wind power, the true value and the predicted value of wind speed are relatively smooth, but the true value of wind power is far from smooth. This shows that the output power rates of two wind turbines with similar wind speed are different even if the turbines are quite close to each other, which fully indicates that the conversion from wind speed to wind power is related to the characteristics of the wind turbine itself. In addition, at the same moment and in the same region, the $MSE$ of wind speed and wind power predicted with the same method are 0.92 and 7.17, respectively. We can see that the $MSE$ of wind speed is much lower than that of wind power. It further shows the wind speed is easier to predict, when the wind power is difficult to predict due to the wind turbine’ specific features. The STF presented in this paper can express wind-related information in a large geographical area and a long time span. Convolutional network can be used to predict the overall variation of the wind in the region and can reduce the effect of "noise" caused by the specific features of the wind turbine. So in the result shown in Figure 8, the predicted value is more stable than the true value.

\subsection{Performance Analysis}
The two convolutional networks proposed in this paper can achieve the end-to-end prediction. And since each pixel point at the output end corresponds to a turbine, the prediction of a scene is actually the prediction of all turbines in parallel. Meanwhile, the convolutional network can make full use of GPU acceleration, so the training time has been greatly shorten. The comparative effect of the time for training the model is shown in the last line of table \ref{tab:compare}. It can be seen that, overall, the training time is qualitatively optimized, which has been shorten by a factor of more than 150, in contrast with that of SVR.
\subsection{MSTF Experiment}
As described in Chapter 3, the STF carrying multiple types of information is called MSTF. Using MSTF can further improve the effect of wind power prediction. This paper uses simple experiments to prove this view but will not discuss it in detail.
As shown in table 2, the $MSE$ of MSTF+FC-CNN compared with that of LF+SVR was reduced by 26.69\% on average and 49.83\% at most. Compared with the $MSE$ of LF+kNN, it decreased by 32.49\% on average and 56.63\% at most. The effect is also better than that of using STF. The average $MSE$ of E2E model and FC-CNN model, both of which use MSTF, in comparison with the model using STF are respectively reduced by 7.08\% and 6.81\%.

\section{Conclusion}
This paper proposes a global feature STF for wind power prediction, and uses convolutional network to predict wind power. Compared with the existing methods, the proposed method greatly optimizes the prediction accuracy and the time cost for training models. In addition, this paper also proposes an approach to fuse various types of data by means of MSTF, which is then proved to be effective in the experiment.

In fact, STF is modeling the spatio-temporal state of wind farm, in which wind turbines play the role of information collectors. The denser the wind turbines are, the more completed the information collected is , so STF is quite suitable to describe the state of a large wind farm. It is worth noting that the STF uses plane to represent the spatial state, which will lose the terrain information, so STF is more suitable for the flat area. In the past several years, offshore wind power has grown rapidly. Thanks to the large scale and flat area of offshore wind farms, STF is naturally ideal for modeling and forecasting offshore wind farms. In future work, this paper will focus on offshore wind farms as the main area of application and further develop the following researches.

\begin{itemize}
\item The way of MSTF’s fusion of multiple types of data will be studied in order to continuously improve the accuracy of prediction.
\item In this paper, two kinds of simple models of convolutional network are constructed to make a prediction, and good prediction results have been obtained. In fact, convolutional networks have developed rapidly in recent years, and the next step will be to introduce more advanced models.
\end{itemize}


\bibliographystyle{icml2017}
\bibliography{bibliography}

\end{document}